\documentclass[lettersize,journal]{IEEEtran}
\usepackage{amsmath,amsfonts}
\usepackage{array}
\usepackage{textcomp}
\usepackage{stfloats}
\usepackage{cuted}
\usepackage{url}
\usepackage{graphicx}
\usepackage{cite}
\usepackage{booktabs}
\usepackage{multirow}
\usepackage{xcolor}
\usepackage{xspace}
\usepackage{microtype}
\usepackage[hidelinks]{hyperref}
\usepackage{cleveref}

\newcommand{\method}{MotionShift\xspace}

\begin{document}

\title{MotionShift: Environment-Compatible Motion Transfer via Planner-Evaluator Synergistic Adaptation}

\author{Yanchen~Liu$^{*\dagger}$,~Junyao~Gao$^{*\ddagger}$,~Yanan~Sun$^{\S}$,~Wensheng~Chen$^{\parallel}$,~Guangming~Lu$^{\dagger}$,~and~Wenjie~Pei$^{\dagger}$
\thanks{$^{*}$Equal contribution.}
\thanks{(\textit{Corresponding authors: Yanan Sun and Wenjie Pei.})}
\thanks{$^{\dagger}$Yanchen Liu, Guangming Lu, and Wenjie Pei are with the School of Computer Science and Technology, Harbin Institute of Technology, Shenzhen, China (e-mail: wenjiecoder@outlook.com).}
\thanks{$^{\ddagger}$Junyao Gao is with Tongji University, Shanghai, China.}
\thanks{$^{\S}$Yanan Sun is with the Hong Kong University of Science and Technology, Hong Kong, China (e-mail: now.syn@gmail.com).}
\thanks{$^{\parallel}$Wensheng Chen is with the Shenzhen International Graduate School, Tsinghua University, Shenzhen, China.}}

\markboth{IEEE Transactions on Multimedia}%
{Liu \MakeLowercase{\textit{et al.}}: MotionShift: Environment-Compatible Motion Transfer}

\maketitle

\setlength{\stripsep}{0pt}
\begin{strip}
\vspace{-8em}
{\centering
\includegraphics[width=\textwidth]{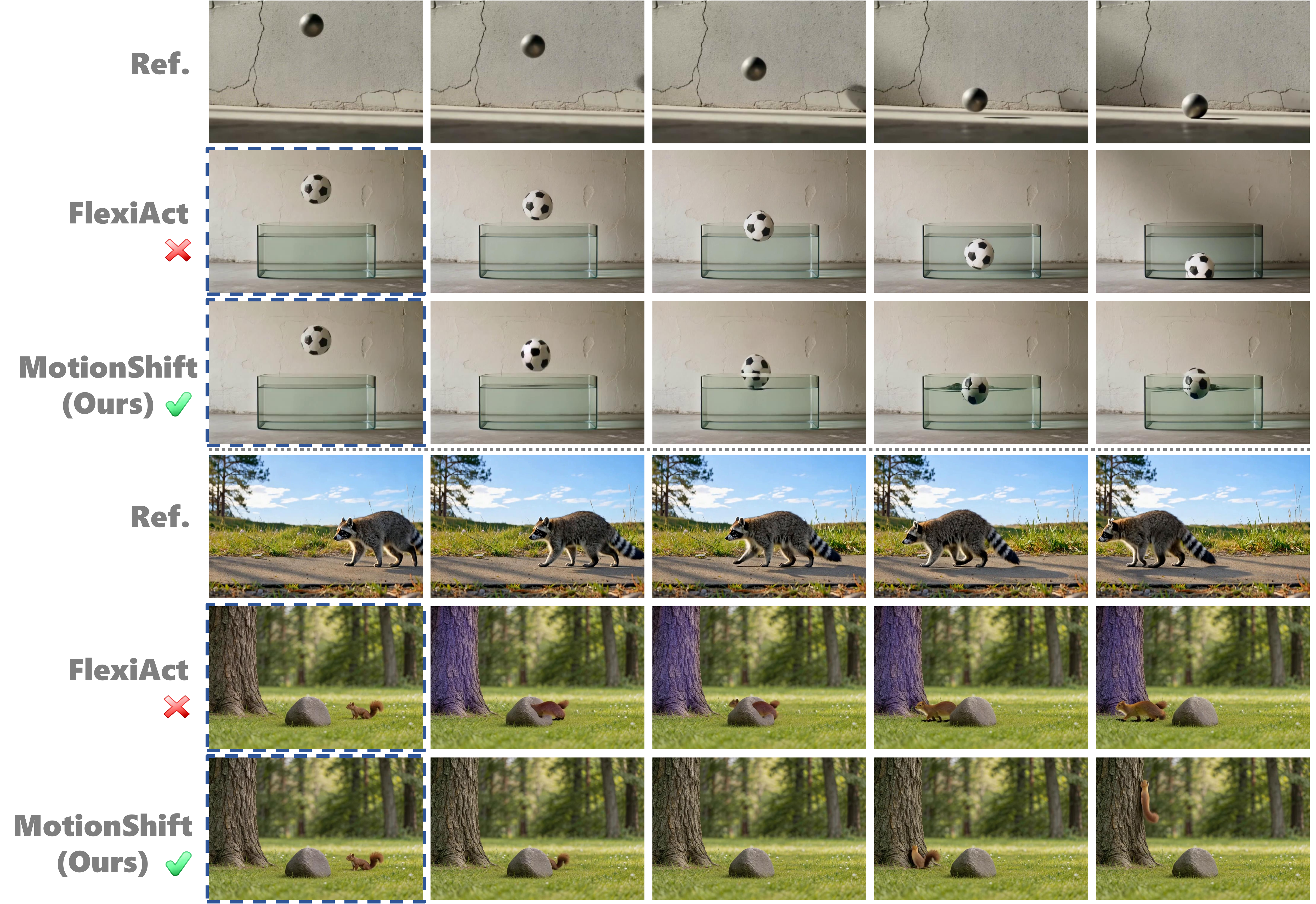}\par}
\refstepcounter{figure}
\label{fig:teaser}
\vspace{4pt}
\small\noindent\textbf{Fig.~\thefigure.}~Visualizations of \textbf{\method} transferring motion from arbitrary reference videos to diverse target scenes (given by the images in blue dashed boxes). Existing methods overlook object-environment interactions, producing physically implausible results. In contrast, MotionShift demonstrates strong adaptability, producing environment-compatible and physically plausible motion results even when complex interactions occur---such as a ball striking the water.
\end{strip}

\begin{abstract}
Existing motion transfer methods excel at exact, rigid motion replication from reference videos to target scenes, whereas they often fail to account for the compatibility between the transferred motion and the target scene environment in accordance with physical laws. In this work, we propose \emph{\method}, a novel motion transfer framework that achieves high-fidelity motion transfer while ensuring physical compatibility with the target scene. In particular, we design a planner--evaluator iterative motion adaptation mechanism, in which a Large Language Model (LLM) acts as the planner to generate motion trajectories by leveraging its strong semantic reasoning capabilities, while a Vision-Language Model (VLM) serves as the evaluator to rectify the motion based on its superior visual perception and scene understanding. Through this planner--evaluator synergistic reasoning process, the motion is adapted to the target scene in a training-free manner. The adapted motion is then used as a generative condition for an image-to-video (I2V) model to render videos in the target scene. Extensive experiments demonstrate that \emph{\method} achieves high-fidelity and physically compatible motion transfer across diverse and complex scenes. Moreover, we show that \emph{\method} can be readily integrated into existing trajectory-conditioned I2V models, endowing them with realistic motion transfer capabilities.
\end{abstract}

\begin{IEEEkeywords}
Motion transfer, video generation, vision-language model, image-to-video synthesis, diffusion model.
\end{IEEEkeywords}

\IEEEPARstart{R}{ecent} advances in diffusion-based image-to-video (I2V) generation~\cite{blattmann2023stable,singer2022make,xing2024dynamicrafter,yang2024cogvideox} have achieved unprecedented visual fidelity and temporal coherence. Building on this foundation, motion transfer techniques have significantly increased controllability by enabling the replication of motion dynamics from reference videos to target images. These methods employ either explicit or implicit motion representations. Explicit approaches leverage interpretable control signals such as depth maps and skeletal keypoints~\cite{guo2024sparsectrl,ma2024follow,wang2023videocomposer,chen2023control} to provide structured motion guidance, while implicit methods~\cite{ling2024motionclone,ren2024customize} encode motion patterns directly from reference videos. More recently, structural alignment methods~\cite{tan2025synmotion,zhang2025flexiact,kansy2025reenact} establish cross-frame object correspondences to propagate spatial layouts.
Despite their effectiveness in motion replication, these techniques largely overlook whether the transferred motion remains physically compatible with the target environment.

Existing motion transfer methods often inject the transferred motion into target scenes without adapting to their spatial structures, physical constraints, or interaction dynamics. This can result in jarring visual conflicts---for instance, a squirrel walking through a rock, as depicted in \cref{fig:teaser}. Such artifacts highlight that current frameworks treat the environment as a passive backdrop rather than an active factor that shapes plausible motion.

These observations reveal an under-explored challenge: \textit{the motion conflicts arising from environmental discrepancies between reference and target inputs}. We formalize this as \textbf{Environment-Compatible Motion Transfer}---the task of transferring motion while ensuring physical plausibility and contextual coherence with the target environment.

In this paper, we propose \textbf{MotionShift}, a motion transfer framework that explicitly addresses environment compatibility. Our framework decomposes the pipeline into two phases: \textbf{Planner--Evaluator Synergistic Motion Adaptation}, which ensures the transferred motion is physically plausible within the target environment, and \textbf{Motion-to-Appearance Video Rendering}, which translates the adapted motion into high-quality video output. In the adaptation phase, we first extract the reference motion trajectory and map it onto the target image via spatial transformation. A planner--evaluator iterative loop then refines this trajectory: the LLM acts as a motion planner that reasons about per-frame object states and outputs corrected bounding boxes, while the VLM serves as a motion evaluator that assesses the corrected trajectory through a pseudo video and provides visual feedback for correction. Through multiple rounds, the trajectory evolves from a coarse positional mapping into an environment-aware motion plan. Notably, this adaptation mechanism is entirely training-free. In the rendering phase, directly generating video from sparse bounding boxes places a heavy burden on a single model to simultaneously resolve motion structure and visual appearance. We instead adopt a progressive strategy: a trained motion generation model first converts the adapted trajectory into a semantic mask sequence that encodes both motion and object-level structure, which is then fed with the target image into an appearance rendering model to produce the final output.

In summary, our main contributions are as follows:
\begin{itemize}
    \item We identify and formalize Environment-Compatible Motion Transfer, a critical yet overlooked challenge in motion transfer for video generation.
    \item We propose a planner--evaluator synergistic motion adaptation strategy that combines semantic reasoning with visual perception to produce environment-compatible motion trajectories in a training-free manner.
    \item We design a progressive motion-to-appearance rendering pipeline that employs semantic motion masks as an intermediate representation, decoupling sparse trajectory control from dense video synthesis.
    \item Extensive experiments show that MotionShift outperforms existing methods in environment compatibility, motion fidelity, and visual quality. We further show that MotionShift can be readily integrated into existing trajectory-conditioned I2V models as a plugin.
\end{itemize}

\section{Related Work}
\label{sec:related}

\subsection{Motion Transfer}
\label{sec:rw_mt}

Recent video diffusion models \cite{blattmann2023stable,brooks2024video,chen2024videocrafter2,singer2022make,yang2024cogvideox} have made remarkable progress in visual fidelity and temporal coherence, yet generating physically plausible object-environment interactions remains an open challenge \cite{liu2024physgen,yuan2025newtongen,escontrela2023video}. This challenge is particularly pronounced in motion transfer, which aims to apply motion patterns from a reference source to a target image. Existing approaches employ either explicit or implicit motion representations. Explicit methods use interpretable control signals such as depth maps \cite{chen2023control,esser2023structure,guo2024sparsectrl,wang2023videocomposer,xing2024make}, skeletal keypoints \cite{hu2024animate,ma2024follow}, optical flow \cite{lei2025animateanything,niu2024mofa,wang2024motionctrl}, or attention maps \cite{gao2025faceshot,meral2024motionflow} to guide motion generation, while implicit methods, including fine-tuning \cite{bi2025customttt,jeong2024vmc,materzynska2024newmove,ren2024customize} and inversion-then-generation paradigms \cite{ling2024motionclone,bai2025uniedit,yatim2024space}, encode motion into model weights or extract motion features from attention maps. However, both types tend to entangle motion with the reference appearance, leading to appearance leakage and poor compatibility with the target scene.

Some works \cite{materzynska2024newmove,wei2024dreamvideo,wu2024lamp,zhao2024motiondirector} mitigate these issues by using multiple reference videos to avoid overfitting to a single layout. Others focus on decoupling motion and appearance to extract semantic-level motion representations \cite{tan2025synmotion,kansy2025reenact,zhang2025flexiact} or reconstructing motion information \cite{liu2025motionshot} to better align with the target object. Nonetheless, these methods generally overlook the compatibility between transferred motion and the target environment, often resulting in physically or logically implausible motions. Our work explicitly addresses this gap by introducing environment-aware motion adaptation before video generation.

\subsection{Large Model Enhanced Video Generation}
\label{sec:rw_lm}

Large language models (LLMs) and vision-language models (VLMs) are increasingly valuable in controllable video generation. LLMs enhance short prompts by providing detailed semantic descriptions \cite{lin2023videodirectorgpt,lian2023llm}, structuring multi-scene narratives \cite{he2025dreamstory,hu2024storyagent}, and translating natural-language instructions into coherent multi-step processes \cite{zhang2024llavanextvideo,xue2025phyt2v}. They improve temporal coherence and multi-object controllability in complex scenarios \cite{chen2023videollm,zhang2023video,xiang2024pandora} through high-level reasoning. However, LLMs rely solely on language and can produce hallucinated spatial relations or incorrect interactions. VLMs address these shortcomings by offering object-centric understanding, scene parsing, and multimodal alignment. They effectively reason about spatial layouts \cite{pan2024vlp,lin2024video}, track objects over time \cite{tan2025omnivideodemocratizingunifiedvideo}, and verify consistency with textual or physical constraints \cite{yang2025vlipp}. Nevertheless, VLMs are less effective at long-horizon motion planning and precise motion control, as their outputs are often descriptive rather than directly actionable for generation.

A few recent works have attempted to combine LLMs and VLMs for video generation. For instance, VideoDirectorGPT~\cite{lin2023videodirectorgpt} uses an LLM to plan multi-scene layouts, while PhyT2V~\cite{xue2025phyt2v} employs LLM-guided iterative refinement to improve physical grounding in text-to-video generation. However, these approaches primarily operate at the scene-level or prompt-level: the LLM generates textual plans or revised descriptions, which are then passed to a video generator in a feedforward manner. They do not perform fine-grained, frame-level trajectory reasoning conditioned on the target environment, nor do they incorporate visual feedback to iteratively verify and correct the spatial plausibility of the planned motion.

Our framework addresses these limitations by assigning complementary roles to the two types of models in a closed-loop process: the LLM serves as the planner, performing frame-level trajectory reasoning through its strong semantic and causal understanding, while the VLM serves as the evaluator, providing visual feedback on the spatial plausibility of the planned motion through pseudo-video assessment. This iterative planner--evaluator loop enables fine-grained, environment-aware motion adaptation that goes beyond the scene-level planning of prior approaches.

\subsection{Two-Stage Video Generation}
\label{sec:rw_ts}

Recent works have explored decomposing video generation into multiple stages to reduce the difficulty of directly synthesizing complex videos in a single step. By introducing an intermediate representation that captures motion structure, the generation process can be factored into a motion prediction stage and a subsequent appearance synthesis stage, allowing each to focus on its respective sub-problem. The choice of intermediate representation involves a key trade-off: optical flow \cite{shi2024motion} provides dense pixel-level motion detail but is sensitive to estimation noise and entangles motion with appearance texture; bounding-box trajectories are compact and semantically interpretable but too sparse to directly guide dense video synthesis; object-level semantic masks \cite{yariv2025through,li2026mask2iv} offer a balance — they are more expressive and structurally informative than sparse boxes, while being more error-tolerant and appearance-agnostic than dense flow.

Our Motion-to-Appearance Video Rendering stage adopts semantic masks as the intermediate representation, but the motivation goes beyond this general trade-off. In our setting, the input to the rendering stage is a bounding-box trajectory produced by the planner--evaluator adaptation, which encodes the \emph{intended} motion layout but carries no appearance or texture information. Directly conditioning a video generation model on such sparse signals places a heavy burden on a single model to simultaneously hallucinate object appearance, resolve inter-frame consistency, and respect the adapted motion layout. By first translating the adapted trajectory into a semantic mask sequence and then rendering the final video conditioned on the masks and the target image, we decouple these two sub-problems: the motion generation model focuses on converting sparse spatial intent into dense structural guidance, while the appearance rendering model focuses on synthesizing realistic video conditioned on that structure. This progressive design is particularly suited to environment-compatible motion transfer, where the adapted trajectory may deviate significantly from any seen motion pattern and thus requires a structured bridge to guide dense generation.

\begin{figure*}[!t]
  \centering
  \includegraphics[width=\textwidth]{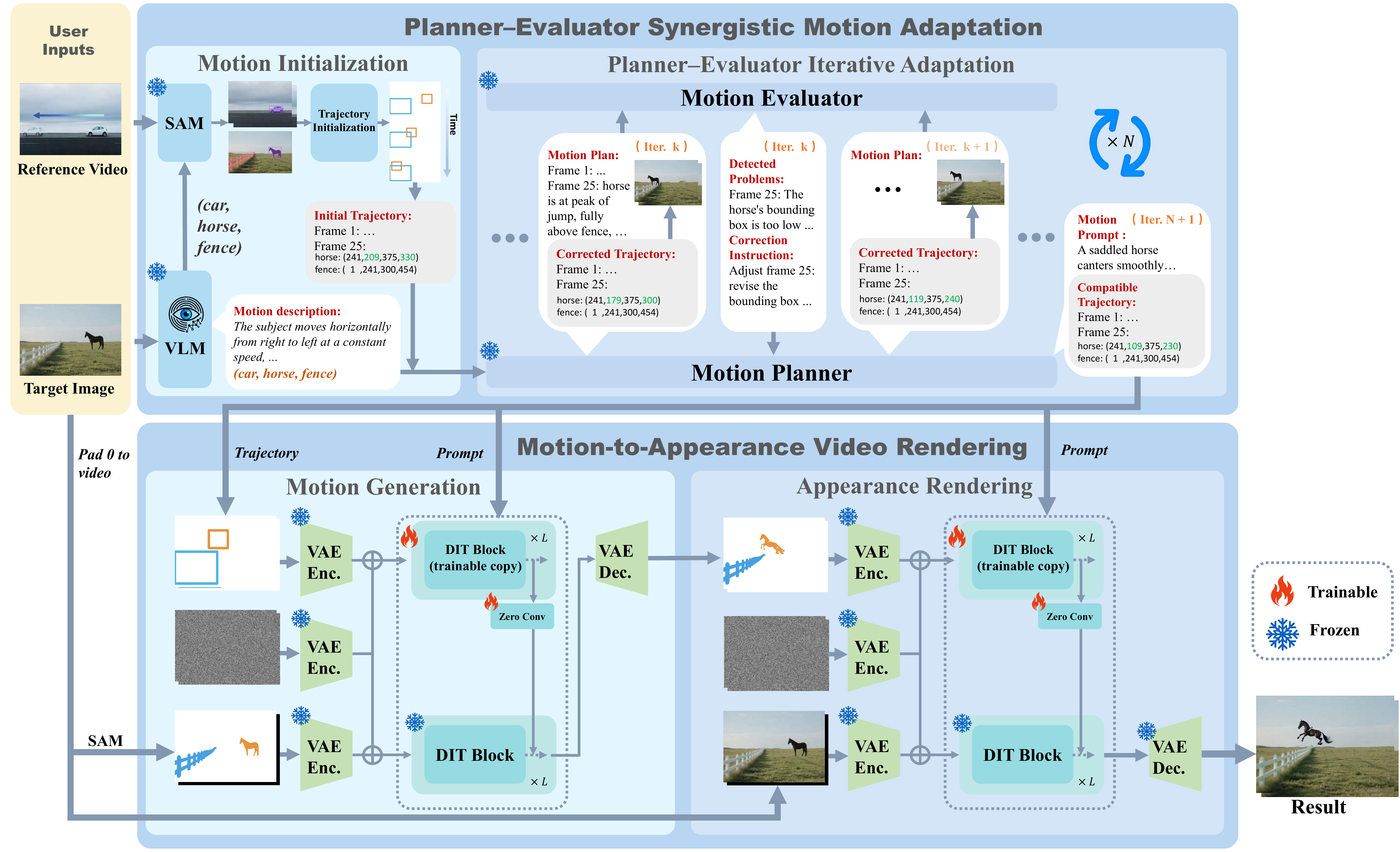}
  \caption{
  \textbf{Overview of \method.} Given a reference video and a target image, we first initialize a motion trajectory for the target scene. Then, through planner--evaluator iterative adaptation, we refine the trajectory to produce an environment-compatible trajectory and a corresponding motion prompt. Finally, these are progressively rendered into the output video: a motion generation model converts the trajectory into a semantic mask sequence, and an appearance rendering model synthesizes the final video from the masks and target image.
  }
  \label{fig:framework}
\end{figure*}

\section{MotionShift}
\label{sec:method}

In this section, we first introduce Environment-Compatible Motion Transfer (ECMT) in \cref{sec:ecmt}. Next, we present the pipeline of MotionShift in \cref{fig:framework} with two key stages: Planner--Evaluator Synergistic Motion Adaptation (\cref{sec:motion_opt}) and Motion-to-Appearance Video Rendering (\cref{sec:motion_render}).

\subsection{Environment-Compatible Motion Transfer}
\label{sec:ecmt}

\begin{figure}[!t]
    \centering
    \includegraphics[width=\linewidth]{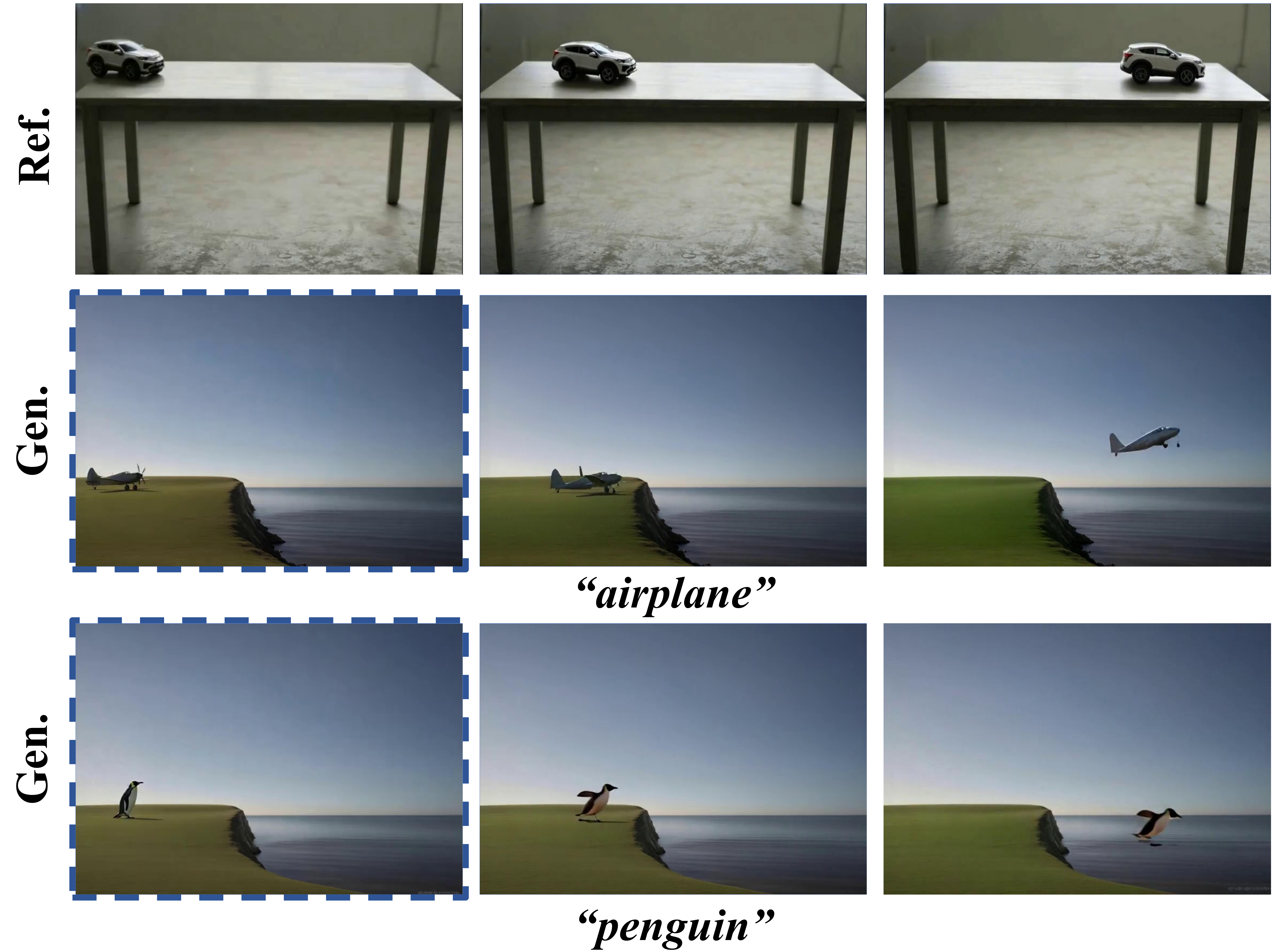}
    \caption{Illustration of environment-compatible motion transfer, where identical reference motions adapt naturally to different scenes, yielding contextually consistent behaviors.}
    \label{fig:ecmt}
\end{figure}

While prior works focus on replicating motion trajectories or temporal patterns, they often overlook whether the transferred motion remains reasonable and coherent within the physical context of the target scene. This limitation leads to visually inconsistent or physically implausible results when the target environment differs significantly from that of the reference video---for instance, when the scene contains distinct terrain, obstacles, or interaction dynamics that alter how the motion should be performed.

To address this fundamental gap, we formalize the task of Environment-Compatible Motion Transfer (ECMT), which explicitly requires that the transferred motion adapt to the spatial structure, semantics, and physical constraints of the target environment. As illustrated in \cref{fig:ecmt}, the same reference motion, when applied to different scenes, may yield divergent yet contextually appropriate motion behaviors, reflecting how environmental conditions naturally influence the motion's evolution. Formally, given a reference video $V_d$ depicting object $O_d$ performing a motion, and a target image $I_s$ containing a corresponding object $O_s$ along with environmental elements $O_i = \{O_i^k\}_{k=1}^{K}$, the goal of ECMT is to generate a video in which $O_s$ performs the transferred motion while maintaining physical plausibility with respect to $O_i$ and the scene geometry---such as respecting gravity, surface continuity, occlusion relationships, and possible interactions with nearby elements.

\subsection{Planner--Evaluator Synergistic Motion Adaptation}
\label{sec:motion_opt}

Achieving environment-compatible motion transfer is challenging, primarily due to the difficulty of collecting high-quality paired videos that share the same motion yet differ in environmental interactions. To address this, we explicitly adopt a two-step strategy: we first initialize a coarse motion trajectory through spatial transformation (\cref{sec:motion_init}), then refine the trajectory via planner--evaluator iterative adaptation to achieve environment compatibility (\cref{sec:vlm_llm_opt}).

\subsubsection{Motion Initialization}
\label{sec:motion_init}

Since dense motion representations such as depth maps cannot be semantically aligned across different object categories, we adopt a sparse bounding-box sequence as the trajectory representation.

We begin by generating an initial target motion trajectory from the reference video and target image. Specifically, we employ a VLM to identify the primary object $O_d$ in $V_d$ and extract a textual motion description $T_d$. We then input $I_s$ along with $T_d$ into the VLM to identify the target object $O_s$ that corresponds to $O_d$, as well as the interacting environmental objects $O_i = \{O_i^k\}_{k=1}^{K}$.

We then utilize SAM~\cite{kirillov2023segment} to segment all identified objects. For the reference video, this yields per-frame bounding boxes forming the reference trajectory:
\begin{equation}
    \mathbf{B}_d = \{\mathbf{b}_d^1, \mathbf{b}_d^2, \ldots, \mathbf{b}_d^F\},
\end{equation}
where $F$ denotes the number of frames. For the target image, SAM provides $\mathbf{b}_s$ for $O_s$ and $\{\mathbf{b}_i^k\}_{k=1}^{K}$ for each $O_i^k$. The motion is then transferred to the target via a relative spatial transformation $\mathcal{T}$, which applies the per-frame displacement and scale change observed in $\mathbf{B}_d$ to the initial position $\mathbf{b}_s$:
\begin{equation}
    \tilde{\mathbf{b}}_s^f = \mathcal{T}(\mathbf{b}_s;\, \mathbf{b}_d^1,\, \mathbf{b}_d^f), \quad f = 1, \ldots, F,
\end{equation}
yielding the initial trajectory $\tilde{\mathbf{B}}_s = \{\tilde{\mathbf{b}}_s^1, \tilde{\mathbf{b}}_s^2, \ldots, \tilde{\mathbf{b}}_s^F\}$ for $O_s$. The bounding boxes of environmental objects $\{\mathbf{b}_i^k\}_{k=1}^{K}$ are replicated across all $F$ frames as static trajectories $\tilde{\mathbf{B}}_i^k$. The complete motion initialization is:
\begin{equation}
    \tilde{\mathbf{B}} = \{\tilde{\mathbf{B}}_s,\, \tilde{\mathbf{B}}_i^1,\, \ldots,\, \tilde{\mathbf{B}}_i^K\}.
\end{equation}

\begin{figure*}[!t]
    \centering
    \includegraphics[width=\textwidth]{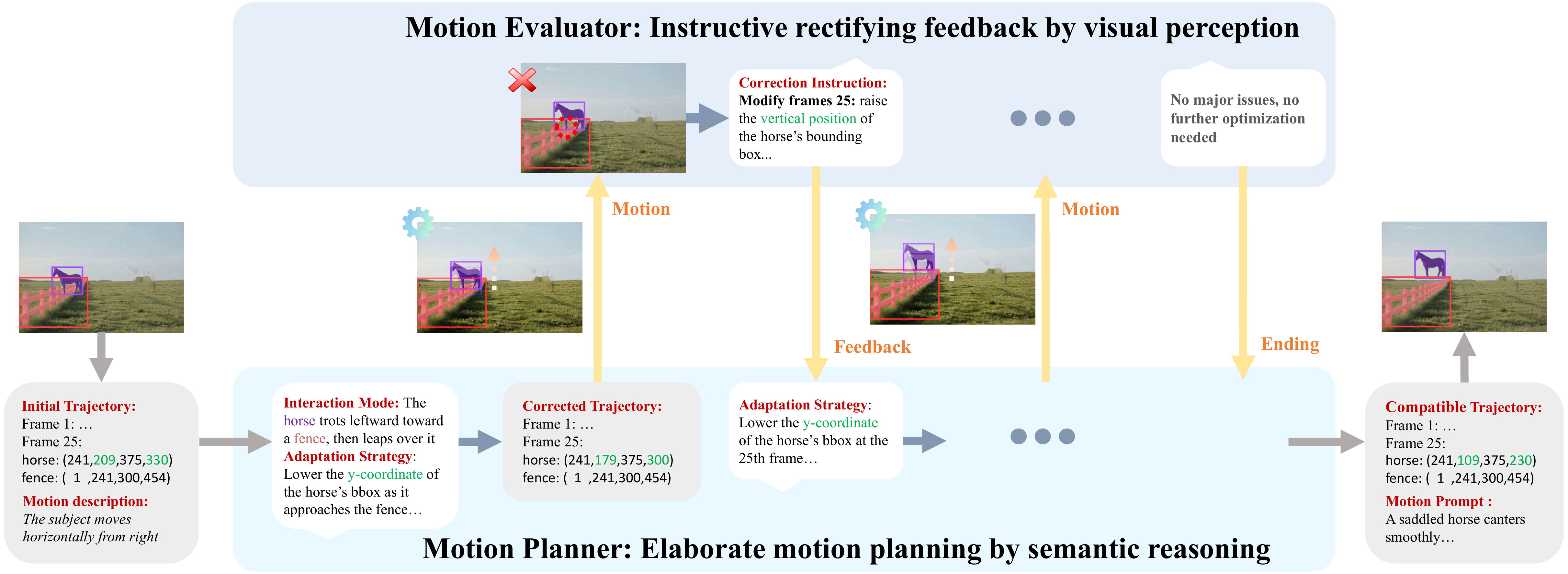}
    \caption{\textbf{Illustration of Planner--Evaluator Iterative Adaptation.} At the beginning of each iteration, we leverage the semantic reasoning capability of the planner (LLM) to infer object--environment interactions and perform detailed motion planning along the given trajectory, producing a corrected motion trajectory. We then construct a pseudo video based on the target image and the corrected trajectory, and feed it to the evaluator (VLM) to assess visual consistency and physical plausibility, which then outputs revision suggestions. These suggestions are then appended to the planner's instruction prompt, and we iteratively refine the motion trajectory until the evaluator detects no major issues.}
    \label{fig:synergistic}
\end{figure*}

\subsubsection{Planner--Evaluator Iterative Adaptation}
\label{sec:vlm_llm_opt}

The initial trajectory merely replicates the reference motion and lacks awareness of object-environment interactions. To obtain an environment-compatible trajectory, we introduce an iterative adaptation loop that leverages the complementary strengths of the planner and evaluator.

\noindent\textbf{Motion Planning.} At each iteration, we provide the planner (LLM) with the current bounding-box trajectory, the motion description $T_d$, and the identities of $O_s$ and $\{O_i^k\}$. The planner is asked to (1) reason about the interaction mode between $O_s$ and environmental objects at each frame (e.g., jumping over an obstacle, being occluded behind a tree, entering water), (2) output corrected bounding boxes $\mathbf{B}_n$ that reflect physically plausible object states, and (3) produce per-frame motion descriptions $\mathbf{P}_n = \{P^{(1)}, P^{(2)}, \ldots, P^{(F)}\}$ that characterize the object's state at each timestep.

\noindent\textbf{Motion Evaluation.} To visually assess the corrected trajectory, we construct a pseudo video from $\mathbf{B}_n$ by masking the objects in $I_s$, inpainting the masked regions~\cite{telea2004inpainting}, and pasting the object crops at the bounding-box positions for each frame. The resulting pseudo video visualizes how the target object moves through the scene over time, making spatial conflicts and implausible object--environment overlaps directly observable. The evaluator (VLM) then evaluates spatial consistency and physical plausibility of the pseudo video. If issues are detected, it outputs correction instructions specifying problematic frames and suggested adjustments, which are appended to the planner's input for the next iteration. The loop terminates when the evaluator reports no major issues remaining.

\noindent\textbf{Motion Summary.} After the adaptation converges, we employ the planner to summarize the per-frame motion descriptions $\mathbf{P}_N$ into a concise video-level motion prompt $T_N$, which serves as the text condition for downstream video generation. This summarization bridges the gap between fine-grained per-frame descriptions and the holistic text condition expected by I2V models.

The final outputs of the Motion Adaptation stage are the environment-compatible trajectory $\mathbf{B}_N$ and the motion prompt $T_N$, as illustrated in \cref{fig:synergistic}.

\subsection{Motion-to-Appearance Video Rendering}
\label{sec:motion_render}

Directly generating video from sparse bounding boxes places a heavy burden on a single model to simultaneously resolve motion structure and visual appearance. To address this, we decompose the rendering process into two stages. In the first stage, a Motion Generation model converts the adapted bounding-box trajectory into a semantic mask sequence, which encodes object-level shape and spatial extent while abstracting away appearance details. In the second stage, an Appearance Rendering model takes the generated mask sequence together with the target image to produce the final video. This progressive design allows each stage to focus on its respective sub-problem.

\subsubsection{Motion Generation}
\label{sec:motion_sem}

The Motion Generation model produces mask-based motion sequences $\hat{\mathbf{M}} = \{\hat{m}^{(1)}, \hat{m}^{(2)}, \ldots, \hat{m}^{(F)}\}$ conditioned on the first-frame segmentation mask $m^{(0)}$, the bounding-box trajectory $\mathbf{B}_N$, and the motion prompt $T_N$. The model is built on CogVideoX-I2V \cite{yang2024cogvideox} with an additional trainable copy branch in a ControlNet-like \cite{zhang2023adding} architecture: the original backbone remains frozen, while the copy branch is initialized from the backbone weights and trained to inject the trajectory condition. The model is denoted as:
\begin{equation}
    \hat{\mathbf{M}}_\theta = f_\theta(\mathbf{z}_t,\, t,\, \mathcal{E}(m^{(0)}),\, \mathcal{E}(\mathbf{B}_N),\, T_N),
\end{equation}
where $\mathbf{z}_t$ is the noisy mask latent at denoising timestep $t$ and $\mathcal{E}$ is the frozen VAE encoder. The generated masks capture the object-level shape and motion layout, providing structured guidance for the subsequent appearance rendering stage.

\subsubsection{Appearance Rendering}
\label{sec:video_gen}

The Appearance Rendering model generates the final output video from the semantic mask sequence $\hat{\mathbf{M}}$ and the target image $I_s$. It adopts the same CogVideoX-I2V backbone with a trainable copy branch as in Motion Generation. The copy branch is trained to inject the mask video as a motion condition: the mask sequence provides object-level spatial guidance for each frame, while the target image supplies appearance information. The model is denoted as:
\begin{equation}
    \hat{\mathbf{V}}_\phi = g_\phi(\mathbf{z}_t,\, t,\, \mathcal{E}(I_s),\, \mathcal{E}(\hat{\mathbf{M}}),\, T_N),
\end{equation}
where $\hat{\mathbf{V}}$ is the generated video. Since the mask sequence already encodes the adapted motion layout and object identity, this model can focus on appearance synthesis and fine-grained temporal detail, completing the progressive rendering from sparse trajectories to dense pixel output.

\begin{figure*}[!t]
  \centering
  \includegraphics[width=\textwidth]{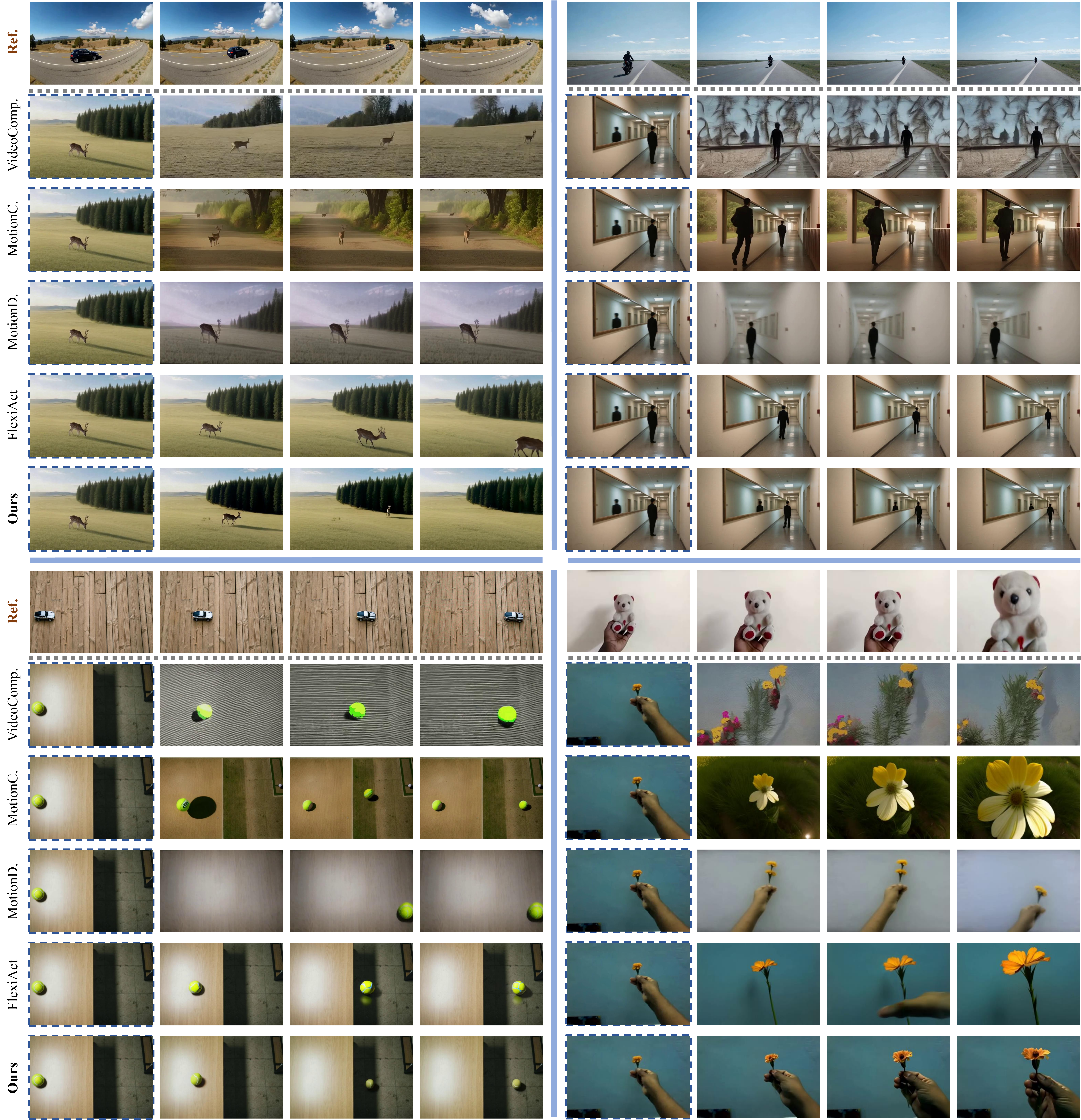}
  \caption{\textbf{Visual comparison with baseline methods.} \method exhibits superior motion consistency and environmental adaptability, accurately capturing perspective variations, realistic reflections, and synchronized hand movements compared to baseline approaches.}
  \label{fig:main_compare}
\end{figure*}

\section{Experiments}
\label{sec:exp}

\subsection{Implementation Details}
\label{sec:impl}

For motion adaptation, we use SAM~\cite{kirillov2023segment} for object segmentation, Qwen-VL-Max~\cite{bai2023qwenvlversatilevisionlanguagemodel} for visual perception and motion evaluation, and o3~\cite{openai2025o3} for motion planning. For video rendering, both the Motion Generation and Appearance Rendering models are built on CogVideoX-5B-I2V~\cite{yang2024cogvideox} with a ControlNet-like trainable copy branch. The original backbone is kept frozen, while the copy branch is initialized from the backbone weights and trained to inject the conditioning signal.

\noindent\textbf{Training Data.} We use MagicData~\cite{li2025magicmotion}, a large-scale trajectory-controlled video dataset comprising 23K video samples, each annotated with a $\langle$video, text, trajectory$\rangle$ triplet including both bounding-box and segmentation mask annotations. For the Motion Generation model, each training sample consists of a first-frame segmentation mask, a bounding-box trajectory sequence, and a ground-truth mask video. For the Appearance Rendering model, each sample consists of a first-frame RGB image, a mask video sequence, and a ground-truth RGB video. All videos are resized to 49 frames at $480 \times 720$ resolution during training.

\noindent\textbf{Optimization.} Both models are trained for one epoch using AdamW optimizer with a learning rate of $1 \times 10^{-5}$ and a batch size of 1 per GPU. Training is conducted on 4 NVIDIA A800 GPUs with gradient accumulation. We use the velocity prediction objective following CogVideoX~\cite{yang2024cogvideox}. Training takes approximately one day per model. During inference, generating a single video takes about 7 minutes on one A800 GPU.

\noindent\textbf{Benchmark.} We use the Something-Something V2 dataset~\cite{goyal2017something}, which contains 174 motion types and 220,847 videos, as our evaluation benchmark following~\cite{kansy2025reenact}. We carefully select 10 diverse motion types from the validation set, each containing 11 videos. For each motion type, one video is randomly selected as the reference video, and the first frame of the remaining ten videos serves as a target image, resulting in 100 test pairs. Every test pair has distinct objects and environmental contexts, making motion transfer particularly challenging. Importantly, the selected action classes involve clear physical interactions between objects and their surroundings --- such as rolling, sliding, and pushing --- where the plausibility of the transferred motion depends critically on the spatial structure and physical properties of the target scene. This makes the benchmark well-suited for evaluating environment-compatible motion transfer.

\noindent\textbf{Evaluation Metrics.} Following~\cite{kansy2025reenact,ling2024motionclone,zhang2025flexiact}, we adopt four metrics for automatic evaluation:
\textit{(1) Appearance consistency}: we compute the CLIP-score~\cite{radford2021learning} between each frame of the generated video and the target image, then average over all frames. This reflects how well the generated video preserves the visual identity of the target subject.
\textit{(2) Temporal consistency}: we compute the CLIP-score between every pair of consecutive frames and average over all pairs, capturing the stability of appearance over time.
\textit{(3) Motion accuracy}: following~\cite{tan2025synmotion}, we use a VLM (InternVL2~\cite{chen2023internvl}) to judge whether the generated video depicts the same motion pattern as the reference video, querying each video ten times and reporting the mean accuracy.
\textit{(4) Motion appropriateness}: we use InternVL2 to assess whether the generated motion appears natural, smooth, and physically plausible. Each video is evaluated ten times, and we report the mean score.
Additionally, we conduct a user study where 20 volunteers rate generated videos on a 5-point scale across all four dimensions. For each dimension, we randomly select ten examples from the test set and present the generated videos from all methods side by side, reporting the average rating across all volunteers and examples.

\begin{table*}[!t]
\centering
\caption{Quantitative comparison. Our method significantly outperforms the other leading methods.}
\label{tab:quantitative}
\renewcommand{\arraystretch}{1.2}
\resizebox{\textwidth}{!}{
\begin{tabular}{l|cccc|cccc}
\toprule
\multirow{3}{*}{Method} & \multicolumn{4}{c|}{Automatic Evaluation $\uparrow$} & \multicolumn{4}{c}{User Study $\uparrow$} \\
\cmidrule(lr){2-5} \cmidrule(lr){6-9}
& Appearance & Temporal & Motion & Motion & Appearance & Temporal & Motion & Motion \\
& Consistency & Consistency & Accuracy & Appropriateness & Consistency & Consistency & Accuracy & Appropriateness \\
\midrule
VideoComposer~\cite{wang2023videocomposer} & 73.44 & 92.77 & 18.60 & 26.70 & 2.15 & 2.23 & 2.08 & 2.19 \\
MotionClone~\cite{ling2024motionclone} & 71.25 & 94.57 & 11.60 & 32.20 & 2.31 & 2.38 & 2.24 & 2.27 \\
MotionDirector~\cite{zhao2024motiondirector} & 79.78 & 98.56 & 10.90 & 52.80 & 2.18 & 2.26 & 2.14 & 2.11 \\
FlexiAct~\cite{zhang2025flexiact} & 85.68 & 97.77 & 30.10 & 77.60 & 3.52 & 3.61 & 3.48 & 3.57 \\
\textbf{Ours} & \textbf{86.53} & \textbf{99.01} & \textbf{53.80} & \textbf{90.40} & \textbf{4.81} & \textbf{4.72} & \textbf{4.75} & \textbf{4.68} \\
\bottomrule
\end{tabular}
}
\end{table*}

\subsection{Comparison with State-of-the-Art Methods}
\label{sec:comparison}

\noindent\textbf{Qualitative Results.} We compare MotionShift with state-of-the-art motion transfer methods, including VideoComposer~\cite{wang2023videocomposer}, MotionClone~\cite{ling2024motionclone}, MotionDirector~\cite{zhao2024motiondirector}, and FlexiAct~\cite{zhang2025flexiact}.
As shown in \cref{fig:main_compare}, VideoComposer and MotionClone exhibit noticeable visual inconsistencies and artifacts in both objects and the surrounding environment, as their dense conditioning tends to rigidly replicate the spatial structure during motion transfer. MotionDirector generates videos with misaligned appearance and motion due to insufficient implicit spatial and temporal fine-tuning. FlexiAct produces visually consistent results but fails to capture object--environment interactions; for example, the mirror does not reflect the walking person and the ball does not fall to the ground. In contrast, MotionShift generates videos with both faithful motion transfer and plausible environment interactions, demonstrating its effectiveness in environment-compatible motion transfer.

\begin{figure}[!t]
    \centering
    \includegraphics[width=\linewidth]{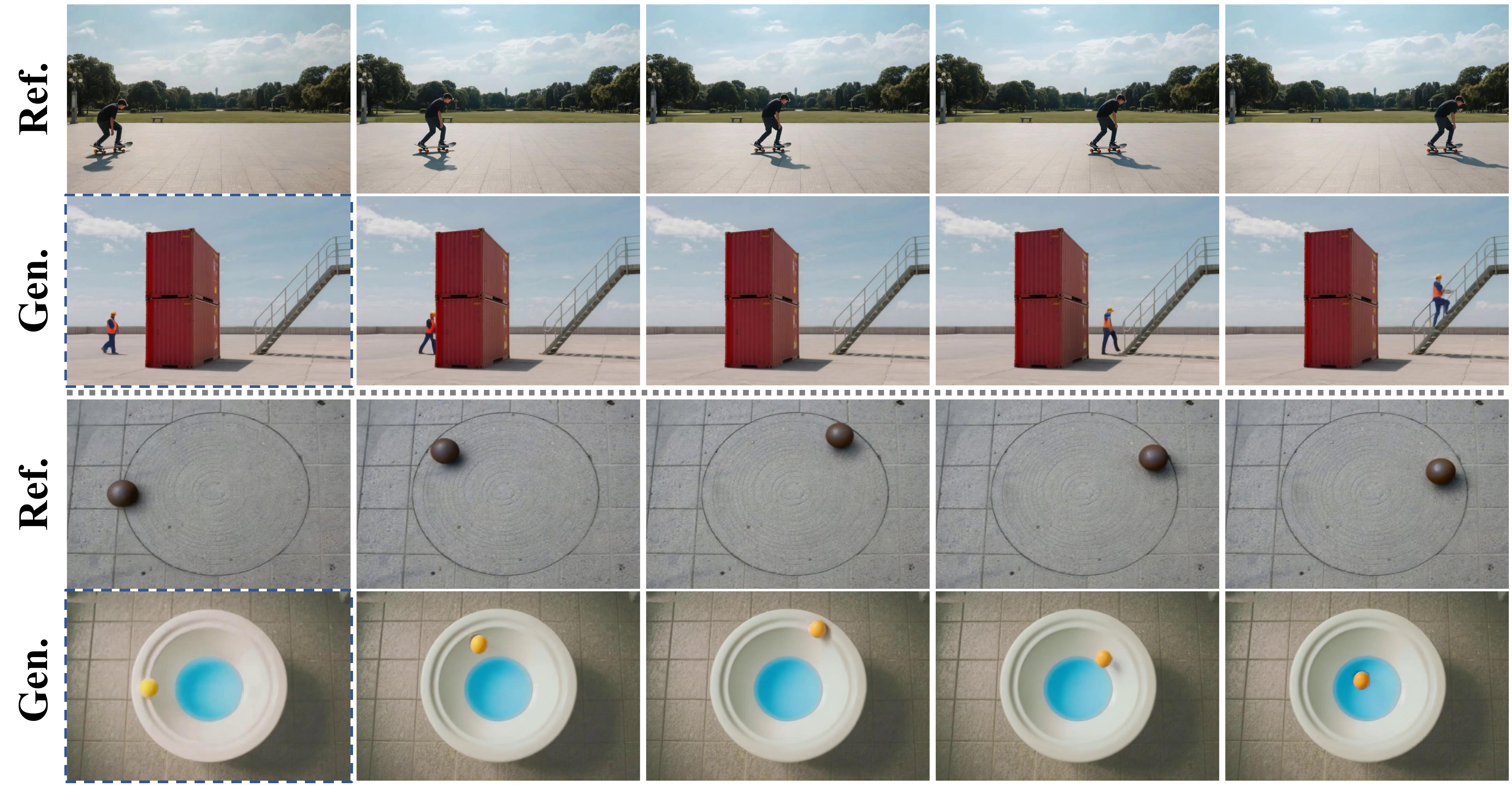}
    \caption{Visualizations of \method in scenes with complex interactions.}
    \label{fig:complex}
\end{figure}

We further demonstrate MotionShift's ability to generalize to complex interactions in \cref{fig:complex}. For example, the worker is naturally occluded by the container and walks up the stairs while preserving the basic motion from the reference video. Moreover, the ball follows circular motion in the reference video but, when transferred to a new scene, first spirals along the bowl wall, then gradually slides down due to gravity, decelerates upon hitting the water, and finally floats on the surface. These results illustrate that MotionShift is capable of handling scenarios involving multi-step physical interactions, where the motion trajectory must be adapted not only to avoid collisions but also to respect continuous physical processes such as gravity, friction, and buoyancy across successive frames.

\noindent\textbf{Quantitative Results.} As shown in \cref{tab:quantitative}, MotionShift achieves the best performance across all metrics, outperforming all baselines in both automatic evaluation and user study. Notably, the performance gap is most pronounced on motion appropriateness: MotionShift scores 90.40 in automatic evaluation, compared to 77.60 for the strongest baseline FlexiAct, and achieves a user study rating of 4.68 out of 5 versus 3.57. This large margin on the environment-compatibility-related metrics, relative to the smaller gap on appearance and temporal consistency, confirms that MotionShift's advantage comes specifically from its environment-aware trajectory adaptation rather than from improvements in rendering quality alone. The user study ratings across all four dimensions further validate the effectiveness of our approach, with MotionShift receiving consistently high scores from human evaluators.

\subsection{Ablation Study}
\label{sec:ablation}

\begin{figure}[!t]
    \centering
    \includegraphics[width=\linewidth]{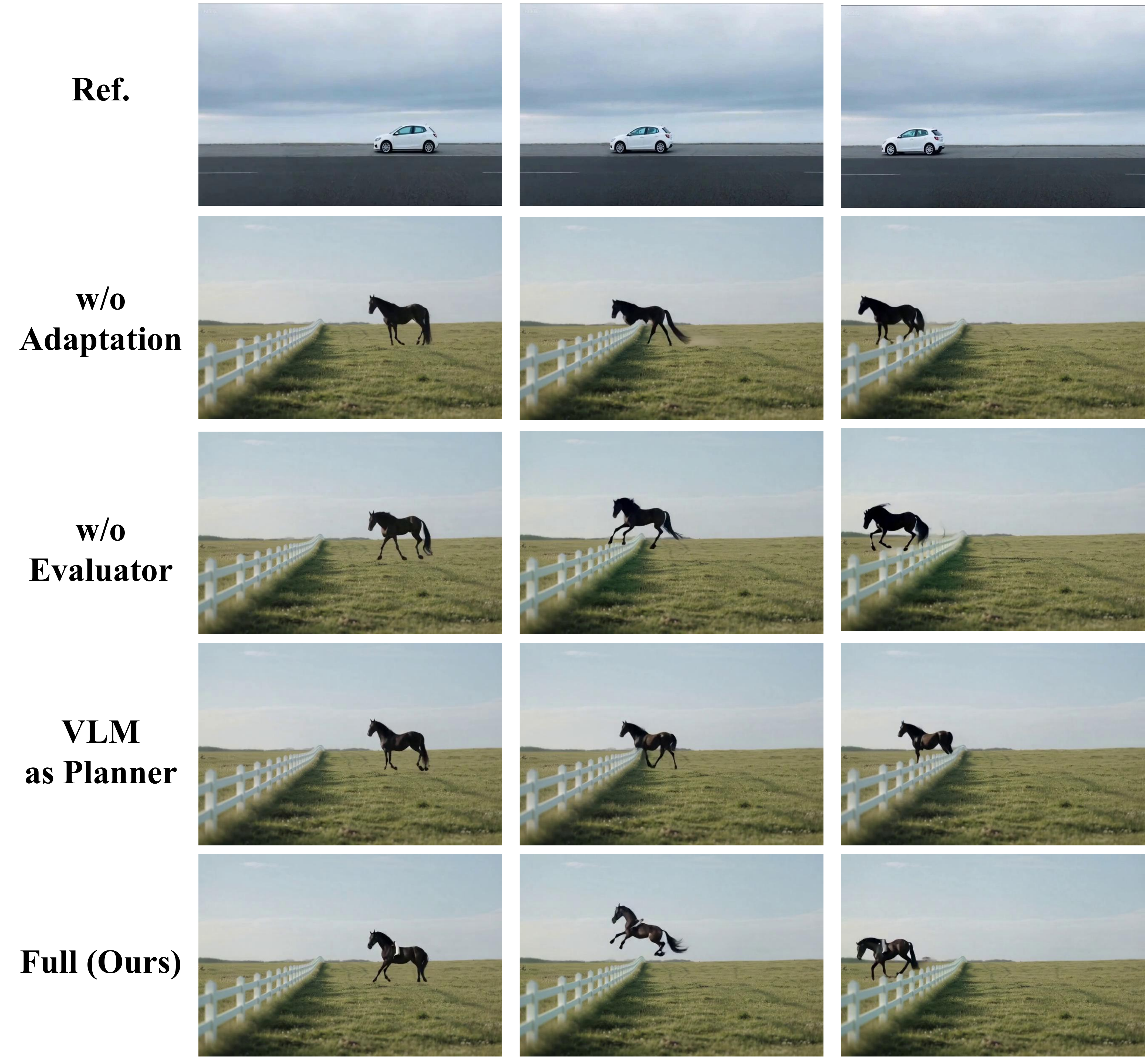}
    \caption{Ablation on planner--evaluator adaptation. We compare variants by disabling adaptation, removing the evaluator, or replacing the planner.}
    \label{fig:ablation_opt}
\vspace{-2em}
\end{figure}

\begin{figure}[!t]
    \centering
    \includegraphics[width=\linewidth]{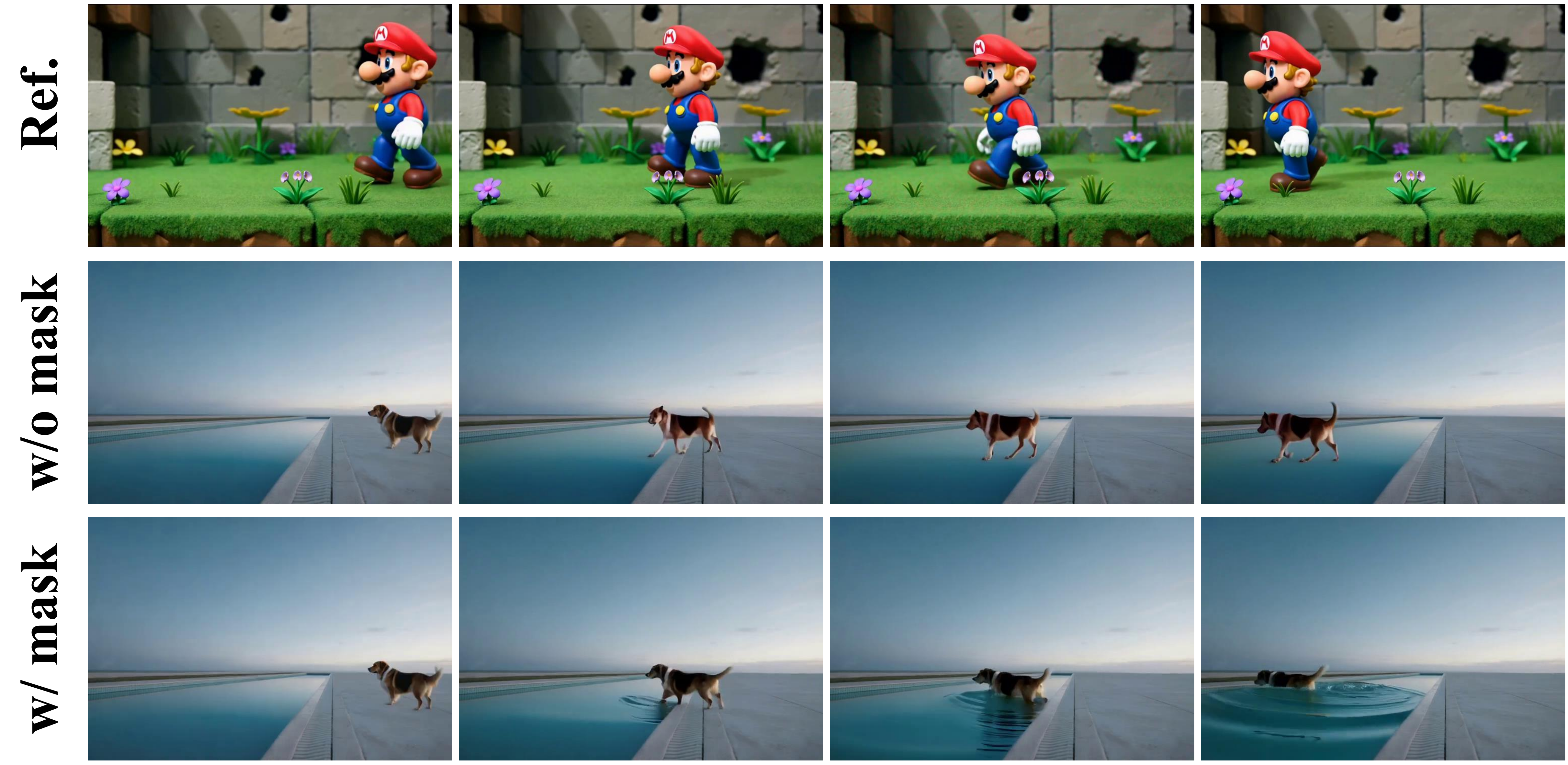}
    \caption{Ablation on motion-to-appearance rendering: w/o mask vs. w/ mask (ours).}
    \label{fig:ablation_render}
\end{figure}

\begin{figure}[!t]
    \centering
    \includegraphics[width=\linewidth]{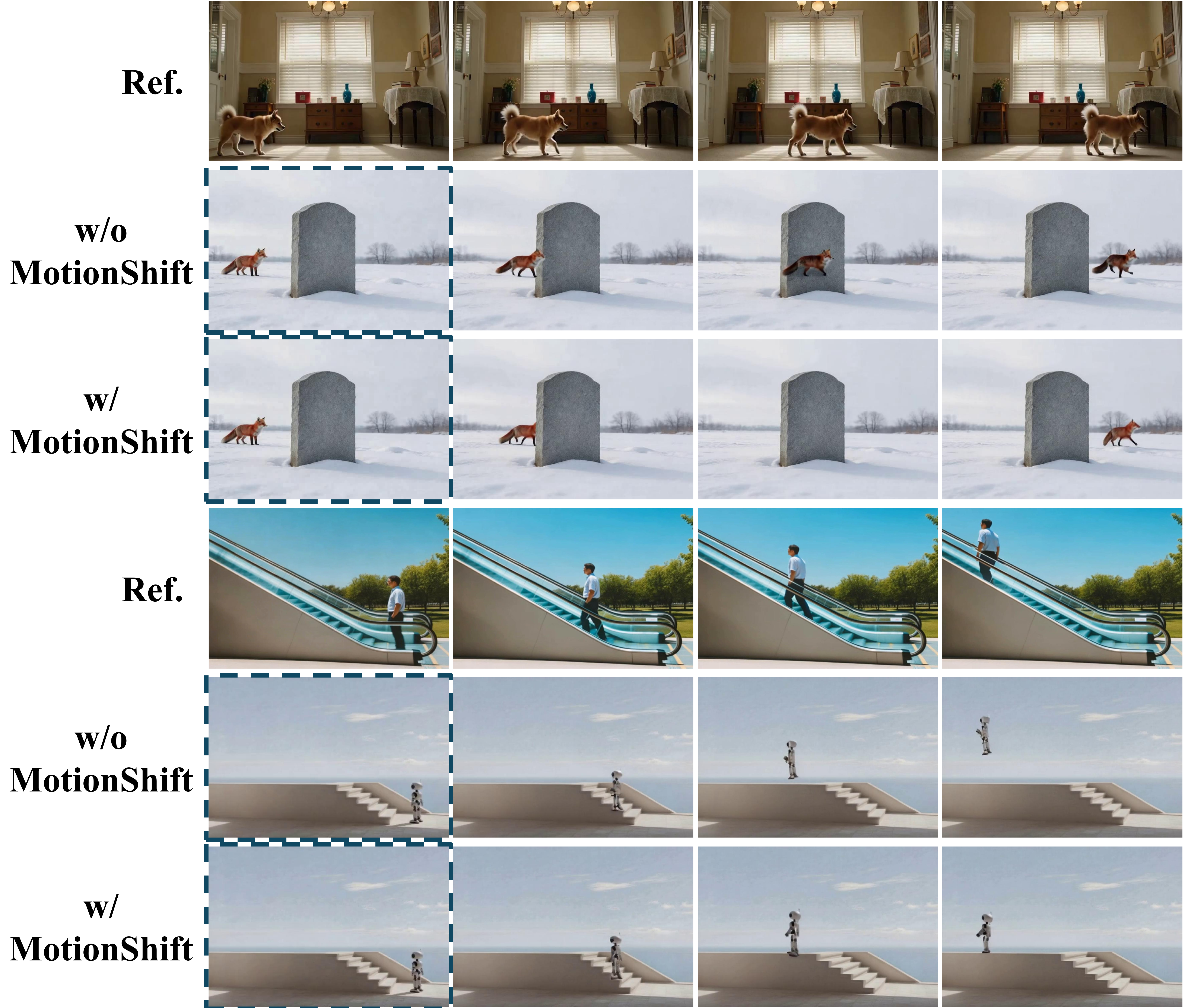}
    \caption{Visual results of Go-with-the-Flow using the initial trajectory vs. the environment-compatible trajectory produced by \method as a plugin.}
    \label{fig:plugin}
\end{figure}

\noindent\textbf{Planner--Evaluator Adaptation.} To verify the effectiveness of our planner--evaluator design, we compare four variants. As shown in \cref{fig:ablation_opt}, without adaptation (i.e., using the initial trajectory directly), the horse collides with the fence, producing noticeable artifacts. Using the planner (LLM) alone without the evaluator successfully reasons that the horse should jump, but without visual feedback, its legs still intersect the fence. Using a VLM instead of an LLM as the planner yields results similar to the initial trajectory, as the VLM can perceive visual conflicts but lacks the long-horizon reasoning ability to infer that the horse should jump. Only by combining both---the LLM as planner for motion reasoning and the VLM as evaluator for visual assessment---does the generated video achieve both visual consistency and environment compatibility.

\noindent\textbf{Motion-to-Appearance Rendering.} We ablate the two-stage rendering design by comparing two variants: (1) generating video directly from the adapted trajectory without semantic masks (w/o mask), and (2) our full pipeline that first generates semantic masks and then renders the final video (w/ mask). As shown in \cref{fig:ablation_render}, without the intermediate mask representation, the generated video suffers from distorted object appearance and incorrect motion behavior, as the sparse trajectory alone provides insufficient guidance for the model to faithfully reconstruct both appearance and dynamics. With semantic masks as an intermediate representation, the object appearance and motion are significantly improved, as the masks provide dense, frame-level structural guidance that bridges sparse trajectory control and high-quality video synthesis.

\subsection{As a Plugin}
\label{sec:plugin}

The adaptation stage of MotionShift outputs an environment-compatible trajectory and a motion prompt, which can be applied to any trajectory-conditioned I2V model without modification. To verify this, we feed the adapted trajectory into the Go-with-the-Flow~\cite{burgert2025go} pipeline. As shown in \cref{fig:plugin}, Go-with-the-Flow produces unnatural videos when following the initial trajectory, such as a robot walking into the sky. When incorporating the environment-compatible trajectory from MotionShift, it produces coherent, natural, and visually harmonious results. This demonstrates that the planner--evaluator adaptation stage is fully decoupled from the rendering backbone and can serve as a general-purpose preprocessing module: since it operates entirely on bounding-box trajectories and natural language, it is agnostic to the choice of downstream I2V model. As a result, any future advancement in trajectory-conditioned video generation can directly benefit from MotionShift's environment-aware adaptation without retraining or architectural changes.

\section{Limitation}

Our method adopts bounding-box sequences as the motion representation, which effectively captures object-level spatial movements but is inherently unable to represent fine-grained actions such as specific limb postures, deformable object shapes, or articulated body movements. As a result, MotionShift is best suited for transferring holistic object-level motion patterns — such as translation, rotation, and trajectory-level dynamics — rather than detailed action semantics that require part-level control. Extending the framework to support richer motion representations, such as keypoint-based or part-aware trajectories, that can capture fine-grained actions while remaining compatible with environment adaptation remains an important direction for future work.

The planner--evaluator adaptation stage also involves multiple rounds of LLM and VLM inference, which incurs non-trivial computational overhead per sample (approximately 7 minutes per video on a single A800 GPU). Additionally, the rendering stage is built upon CogVideoX-5B-I2V as the backbone; while this model delivers competitive quality, stronger base models such as Wan2.1~\cite{wan2025wan} or future successors may further improve the visual fidelity and motion realism of the generated videos without changes to the adaptation mechanism. Exploring more efficient adaptation strategies and integrating stronger base models are promising directions for reducing inference cost and improving output quality.

\section{Conclusion}

In this work, we introduce \method, the first framework to address environment-compatible motion transfer. We identify and formalize this task, which requires that the transferred motion remain physically plausible and contextually coherent with the target scene — a challenge largely overlooked by existing motion transfer methods. To address it, we propose a planner--evaluator synergistic adaptation mechanism that iteratively refines motion trajectories through semantic reasoning from an LLM planner and visual feedback from a VLM evaluator, in a training-free manner. The adapted motion is then progressively rendered into the final video through a two-stage pipeline that uses semantic masks as an intermediate representation, decoupling sparse trajectory control from dense appearance synthesis.

Extensive experiments demonstrate that \method achieves superior environment compatibility, motion fidelity, and visual consistency compared to existing baselines. We further show that \method exhibits strong generalization to scenarios involving complex object-environment interactions, and can be readily integrated into existing trajectory-conditioned I2V models as a plugin, improving their physical plausibility without retraining. We believe this work highlights an important but under-explored dimension of controllable video generation, and opens new research directions at the intersection of environment-aware reasoning and physical video synthesis.

\bibliographystyle{IEEEtran}
\bibliography{main}

@String(CVPR= {IEEE Conf. Comput. Vis. Pattern Recog.})

@String(ICCV= {Int. Conf. Comput. Vis.})

@String(ECCV= {Eur. Conf. Comput. Vis.})

@String(TVCG  = {IEEE Trans. Vis. Comput. Graph.})

@String(ACCV  = {ACCV})

@String(AAAI = {AAAI})

@String(CVPR  = {CVPR})

@String(ICCV  = {ICCV})

@String(ECCV  = {ECCV})

@String(TVCG  = {IEEE TVCG})

@article{blattmann2023stable,
  title={Stable video diffusion: Scaling latent video diffusion models to large datasets},
  author={Blattmann, Andreas and Dockhorn, Tim and Kulal, Sumith and Mendelevitch, Daniel and Kilian, Maciej and Lorenz, Dominik and Levi, Yam and English, Zion and Voleti, Vikram and Letts, Adam and others},
  journal={arXiv preprint arXiv:2311.15127},
  year={2023}
}

@article{singer2022make,
  title={Make-a-video: Text-to-video generation without text-video data},
  author={Singer, Uriel and Polyak, Adam and Hayes, Thomas and Yin, Xi and An, Jie and Zhang, Songyang and Hu, Qiyuan and Yang, Harry and Ashual, Oron and Gafni, Oran and others},
  journal={arXiv preprint arXiv:2209.14792},
  year={2022}
}

@inproceedings{xing2024dynamicrafter,
  title={Dynamicrafter: Animating open-domain images with video diffusion priors},
  author={Xing, Jinbo and Xia, Menghan and Zhang, Yong and Chen, Haoxin and Yu, Wangbo and Liu, Hanyuan and Liu, Gongye and Wang, Xintao and Shan, Ying and Wong, Tien-Tsin},
  booktitle={ECCV},
  pages={399--417},
  year={2024},
  organization={Springer}
}

@article{yang2024cogvideox,
  title={Cogvideox: Text-to-video diffusion models with an expert transformer},
  author={Yang, Zhuoyi and Teng, Jiayan and Zheng, Wendi and Ding, Ming and Huang, Shiyu and Xu, Jiazheng and Yang, Yuanming and Hong, Wenyi and Zhang, Xiaohan and Feng, Guanyu and others},
  journal={arXiv preprint arXiv:2408.06072},
  year={2024}
}

@inproceedings{zhao2024motiondirector,
  title={Motiondirector: Motion customization of text-to-video diffusion models},
  author={Zhao, Rui and Gu, Yuchao and Wu, Jay Zhangjie and Zhang, David Junhao and Liu, Jia-Wei and Wu, Weijia and Keppo, Jussi and Shou, Mike Zheng},
  booktitle={ECCV},
  pages={273--290},
  year={2024},
  organization={Springer}
}

@article{chen2023control,
  title={Control-a-video: Controllable text-to-video generation with diffusion models},
  author={Chen, Weifeng and Ji, Yatai and Wu, Jie and Wu, Hefeng and Xie, Pan and Li, Jiashi and Xia, Xin and Xiao, Xuefeng and Lin, Liang},
  journal={arXiv e-prints},
  pages={arXiv--2305},
  year={2023}
}

@inproceedings{esser2023structure,
  title={Structure and content-guided video synthesis with diffusion models},
  author={Esser, Patrick and Chiu, Johnathan and Atighehchian, Parmida and Granskog, Jonathan and Germanidis, Anastasis},
  booktitle={ICCV},
  pages={7346--7356},
  year={2023}
}

@inproceedings{guo2024sparsectrl,
  title={Sparsectrl: Adding sparse controls to text-to-video diffusion models},
  author={Guo, Yuwei and Yang, Ceyuan and Rao, Anyi and Agrawala, Maneesh and Lin, Dahua and Dai, Bo},
  booktitle={ECCV},
  pages={330--348},
  year={2024},
  organization={Springer}
}

@article{wang2023videocomposer,
  title={Videocomposer: Compositional video synthesis with motion controllability},
  author={Wang, Xiang and Yuan, Hangjie and Zhang, Shiwei and Chen, Dayou and Wang, Jiuniu and Zhang, Yingya and Shen, Yujun and Zhao, Deli and Zhou, Jingren},
  journal={NeurIPS},
  volume={36},
  pages={7594--7611},
  year={2023}
}

@article{xing2024make,
  title={Make-your-video: Customized video generation using textual and structural guidance},
  author={Xing, Jinbo and Xia, Menghan and Liu, Yuxin and Zhang, Yuechen and Zhang, Yong and He, Yingqing and Liu, Hanyuan and Chen, Haoxin and Cun, Xiaodong and Wang, Xintao and others},
  journal={TVCG},
  volume={31},
  number={2},
  pages={1526--1541},
  year={2024},
  publisher={IEEE}
}

@inproceedings{hu2024animate,
  title={Animate anyone: Consistent and controllable image-to-video synthesis for character animation},
  author={Hu, Li},
  booktitle={CVPR},
  pages={8153--8163},
  year={2024}
}

@inproceedings{ma2024follow,
  title={Follow your pose: Pose-guided text-to-video generation using pose-free videos},
  author={Ma, Yue and He, Yingqing and Cun, Xiaodong and Wang, Xintao and Chen, Siran and Li, Xiu and Chen, Qifeng},
  booktitle={AAAI},
  volume={38},
  pages={4117--4125},
  year={2024}
}

@article{gao2025faceshot,
  title={Faceshot: Bring any character into life},
  author={Gao, Junyao and Sun, Yanan and Shen, Fei and Jiang, Xin and Xing, Zhening and Chen, Kai and Zhao, Cairong},
  journal={arXiv preprint arXiv:2503.00740},
  year={2025}
}

@article{ling2024motionclone,
  title={Motionclone: Training-free motion cloning for controllable video generation},
  author={Ling, Pengyang and Bu, Jiazi and Zhang, Pan and Dong, Xiaoyi and Zang, Yuhang and Wu, Tong and Chen, Huaian and Wang, Jiaqi and Jin, Yi},
  journal={arXiv preprint arXiv:2406.05338},
  year={2024}
}

@article{meral2024motionflow,
  title={Motionflow: Attention-driven motion transfer in video diffusion models},
  author={Meral, Tuna Han Salih and Yesiltepe, Hidir and Dunlop, Connor and Yanardag, Pinar},
  journal={arXiv preprint arXiv:2412.05275},
  year={2024}
}

@inproceedings{bi2025customttt,
  title={Customttt: Motion and appearance customized video generation via test-time training},
  author={Bi, Xiuli and Lu, Jian and Liu, Bo and Cun, Xiaodong and Zhang, Yong and Li, Weisheng and Xiao, Bin},
  booktitle={AAAI},
  volume={39},
  pages={1871--1879},
  year={2025}
}

@inproceedings{jeong2024vmc,
  title={Vmc: Video motion customization using temporal attention adaption for text-to-video diffusion models},
  author={Jeong, Hyeonho and Park, Geon Yeong and Ye, Jong Chul},
  booktitle={CVPR},
  pages={9212--9221},
  year={2024}
}

@inproceedings{materzynska2024newmove,
  title={Newmove: Customizing text-to-video models with novel motions},
  author={Materzy{\'n}ska, Joanna and Sivic, Josef and Shechtman, Eli and Torralba, Antonio and Zhang, Richard and Russell, Bryan},
  booktitle={ACCV},
  pages={1634--1651},
  year={2024}
}

@inproceedings{ren2024customize,
  title={Customize-a-video: One-shot motion customization of text-to-video diffusion models},
  author={Ren, Yixuan and Zhou, Yang and Yang, Jimei and Shi, Jing and Liu, Difan and Liu, Feng and Kwon, Mingi and Shrivastava, Abhinav},
  booktitle={ECCV},
  pages={332--349},
  year={2024},
  organization={Springer}
}

@article{tan2025synmotion,
  title={SynMotion: Semantic-Visual Adaptation for Motion Customized Video Generation},
  author={Tan, Shuai and Gong, Biao and Wei, Yujie and Zhang, Shiwei and Liu, Zhuoxin and Zheng, Dandan and Chen, Jingdong and Wang, Yan and Ouyang, Hao and Zheng, Kecheng and others},
  journal={arXiv preprint arXiv:2506.23690},
  year={2025}
}

@inproceedings{kansy2025reenact,
  title={Reenact anything: Semantic video motion transfer using motion-textual inversion},
  author={Kansy, Manuel and Naruniec, Jacek and Schroers, Christopher and Gross, Markus and Weber, Romann M},
  booktitle={SIGGRAPH},
  pages={1--12},
  year={2025}
}

@inproceedings{zhang2025flexiact,
  title={Flexiact: Towards flexible action control in heterogeneous scenarios},
  author={Zhang, Shiyi and Zhuang, Junhao and Zhang, Zhaoyang and Shan, Ying and Tang, Yansong},
  booktitle={SIGGRAPH},
  pages={1--11},
  year={2025}
}

@inproceedings{chen2024videocrafter2,
  title={Videocrafter2: Overcoming data limitations for high-quality video diffusion models},
  author={Chen, Haoxin and Zhang, Yong and Cun, Xiaodong and Xia, Menghan and Wang, Xintao and Weng, Chao and Shan, Ying},
  booktitle={CVPR},
  pages={7310--7320},
  year={2024}
}

@article{brooks2024video,
  title={Video generation models as world simulators},
  author={Brooks, Tim and Peebles, Bill and Holmes, Connor and DePue, Will and Guo, Yufei and Jing, Li and Schnurr, David and Taylor, Joe and Luhman, Troy and Luhman, Eric and others},
  journal={OpenAI Blog},
  volume={1},
  number={8},
  pages={1},
  year={2024}
}

@article{escontrela2023video,
  title={Video prediction models as rewards for reinforcement learning},
  author={Escontrela, Alejandro and Adeniji, Ademi and Yan, Wilson and Jain, Ajay and Peng, Xue Bin and Goldberg, Ken and Lee, Youngwoon and Hafner, Danijar and Abbeel, Pieter},
  journal={NeurIPS},
  volume={36},
  pages={68760--68783},
  year={2023}
}

@inproceedings{lei2025animateanything,
  title={Animateanything: Consistent and controllable animation for video generation},
  author={Lei, Guojun and Wang, Chi and Zhang, Rong and Wang, Yikai and Li, Hong and Xu, Weiwei},
  booktitle={CVPR},
  pages={27946--27956},
  year={2025}
}

@inproceedings{niu2024mofa,
  title={Mofa-video: Controllable image animation via generative motion field adaptions in frozen image-to-video diffusion model},
  author={Niu, Muyao and Cun, Xiaodong and Wang, Xintao and Zhang, Yong and Shan, Ying and Zheng, Yinqiang},
  booktitle={ECCV},
  pages={111--128},
  year={2024},
  organization={Springer}
}

@inproceedings{wang2024motionctrl,
  title={Motionctrl: A unified and flexible motion controller for video generation},
  author={Wang, Zhouxia and Yuan, Ziyang and Wang, Xintao and Li, Yaowei and Chen, Tianshui and Xia, Menghan and Luo, Ping and Shan, Ying},
  booktitle={SIGGRAPH},
  pages={1--11},
  year={2024}
}

@inproceedings{bai2025uniedit,
  title={Uniedit: A unified tuning-free framework for video motion and appearance editing},
  author={Bai, Jianhong and He, Tianyu and Wang, Yuchi and Guo, Junliang and Hu, Haoji and Liu, Zuozhu and Bian, Jiang},
  booktitle={ACM Multimedia},
  pages={10171--10180},
  year={2025}
}

@inproceedings{yatim2024space,
  title={Space-time diffusion features for zero-shot text-driven motion transfer},
  author={Yatim, Danah and Fridman, Rafail and Bar-Tal, Omer and Kasten, Yoni and Dekel, Tali},
  booktitle={CVPR},
  pages={8466--8476},
  year={2024}
}

@inproceedings{wei2024dreamvideo,
  title={Dreamvideo: Composing your dream videos with customized subject and motion},
  author={Wei, Yujie and Zhang, Shiwei and Qing, Zhiwu and Yuan, Hangjie and Liu, Zhiheng and Liu, Yu and Zhang, Yingya and Zhou, Jingren and Shan, Hongming},
  booktitle={CVPR},
  pages={6537--6549},
  year={2024}
}

@inproceedings{wu2024lamp,
  title={Lamp: Learn a motion pattern for few-shot video generation},
  author={Wu, Ruiqi and Chen, Liangyu and Yang, Tong and Guo, Chunle and Li, Chongyi and Zhang, Xiangyu},
  booktitle={CVPR},
  pages={7089--7098},
  year={2024}
}

@inproceedings{liu2025motionshot,
  title={MotionShot: Adaptive Motion Transfer across Arbitrary Objects for Text-to-Video Generation},
  author={Liu, Yanchen and Sun, Yanan and Xing, Zhening and Gao, Junyao and Chen, Kai and Pei, Wenjie},
  booktitle={ICCV},
  pages={11861--11871},
  year={2025}
}

@inproceedings{pan2024vlp,
  title={Vlp: Vision language planning for autonomous driving},
  author={Pan, Chenbin and Yaman, Burhaneddin and Nesti, Tommaso and Mallik, Abhirup and Allievi, Alessandro G and Velipasalar, Senem and Ren, Liu},
  booktitle={CVPR},
  pages={14760--14769},
  year={2024}
}

@misc{zhang2024llavanextvideo,
  title={LLaVA-NeXT: A Strong Zero-shot Video Understanding Model},
  url={https://llava-vl.github.io/blog/2024-04-30-llava-next-video/},
  author={Zhang, Yuanhan and Li, Bo and Liu, haotian and Lee, Yong jae and Gui, Liangke and Fu, Di and Feng, Jiashi and Liu, Ziwei and Li, Chunyuan},
  month={April},
  year={2024}
}

@inproceedings{xue2025phyt2v,
  title={Phyt2v: Llm-guided iterative self-refinement for physics-grounded text-to-video generation},
  author={Xue, Qiyao and Yin, Xiangyu and Yang, Boyuan and Gao, Wei},
  booktitle={CVPR},
  pages={18826--18836},
  year={2025}
}

@article{lian2023llm,
  title={Llm-grounded video diffusion models},
  author={Lian, Long and Shi, Baifeng and Yala, Adam and Darrell, Trevor and Li, Boyi},
  journal={arXiv preprint arXiv:2309.17444},
  year={2023}
}

@article{lin2023videodirectorgpt,
  title={Videodirectorgpt: Consistent multi-scene video generation via llm-guided planning},
  author={Lin, Han and Zala, Abhay and Cho, Jaemin and Bansal, Mohit},
  journal={arXiv preprint arXiv:2309.15091},
  year={2023}
}

@article{xiang2024pandora,
  title={Pandora: Towards general world model with natural language actions and video states},
  author={Xiang, Jiannan and Liu, Guangyi and Gu, Yi and Gao, Qiyue and Ning, Yuting and Zha, Yuheng and Feng, Zeyu and Tao, Tianhua and Hao, Shibo and Shi, Yemin and others},
  journal={arXiv preprint arXiv:2406.09455},
  year={2024}
}

@article{yang2025vlipp,
  title={VLIPP: Towards Physically Plausible Video Generation with Vision and Language Informed Physical Prior},
  author={Yang, Xindi and Li, Baolu and Zhang, Yiming and Yin, Zhenfei and Bai, Lei and Ma, Liqian and Wang, Zhiyong and Cai, Jianfei and Wong, Tien-Tsin and Lu, Huchuan and others},
  journal={arXiv preprint arXiv:2503.23368},
  year={2025}
}

@inproceedings{liu2024physgen,
  title={Physgen: Rigid-body physics-grounded image-to-video generation},
  author={Liu, Shaowei and Ren, Zhongzheng and Gupta, Saurabh and Wang, Shenlong},
  booktitle={ECCV},
  pages={360--378},
  year={2024},
  organization={Springer}
}

@article{yuan2025newtongen,
  title={NewtonGen: Physics-Consistent and Controllable Text-to-Video Generation via Neural Newtonian Dynamics},
  author={Yuan, Yu and Wang, Xijun and Wickremasinghe, Tharindu and Nadir, Zeeshan and Ma, Bole and Chan, Stanley H},
  journal={arXiv preprint arXiv:2509.21309},
  year={2025}
}

@inproceedings{goyal2017something,
  title={The" something something" video database for learning and evaluating visual common sense},
  author={Goyal, Raghav and Ebrahimi Kahou, Samira and Michalski, Vincent and Materzynska, Joanna and Westphal, Susanne and Kim, Heuna and Haenel, Valentin and Fruend, Ingo and Yianilos, Peter and Mueller-Freitag, Moritz and others},
  booktitle={ICCV},
  pages={5842--5850},
  year={2017}
}

@article{telea2004inpainting,
  title={An image inpainting technique based on the fast marching method},
  author={Telea, Alexandru},
  journal={Journal of Graphics Tools},
  volume={9},
  number={1},
  pages={23--34},
  year={2004},
  publisher={Taylor \& Francis}
}

@misc{bai2023qwenvlversatilevisionlanguagemodel,
      title={Qwen-VL: A Versatile Vision-Language Model for Understanding, Localization, Text Reading, and Beyond}, 
      author={Jinze Bai and Shuai Bai and Shusheng Yang and Shijie Wang and Sinan Tan and Peng Wang and Junyang Lin and Chang Zhou and Jingren Zhou},
      year={2023},
      eprint={2308.12966},
      archivePrefix={arXiv},
      primaryClass={cs.CV},
      url={https://arxiv.org/abs/2308.12966}, 
}

@inproceedings{kirillov2023segment,
  title={Segment anything},
  author={Kirillov, Alexander and Mintun, Eric and Ravi, Nikhila and Mao, Hanzi and Rolland, Chloe and Gustafson, Laura and Xiao, Tete and Whitehead, Spencer and Berg, Alexander C and Lo, Wan-Yen and others},
  booktitle={ICCV},
  pages={4015--4026},
  year={2023}
}

@inproceedings{burgert2025go,
  title={Go-with-the-flow: Motion-controllable video diffusion models using real-time warped noise},
  author={Burgert, Ryan and Xu, Yuancheng and Xian, Wenqi and Pilarski, Oliver and Clausen, Pascal and He, Mingming and Ma, Li and Deng, Yitong and Li, Lingxiao and Mousavi, Mohsen and others},
  booktitle={CVPR},
  pages={13--23},
  year={2025}
}

@article{li2025magicmotion,
  title={Magicmotion: Controllable video generation with dense-to-sparse trajectory guidance},
  author={Li, Quanhao and Xing, Zhen and Wang, Rui and Zhang, Hui and Dai, Qi and Wu, Zuxuan},
  journal={arXiv preprint arXiv:2503.16421},
  year={2025}
}

@misc{openai2025o3,
  title={Deliberative Alignment in {OpenAI} o3},
  author={{OpenAI}},
  year={2025},
  url={https://openai.com/index/deliberative-alignment/}
}

@inproceedings{radford2021learning,
  title={Learning transferable visual models from natural language supervision},
  author={Radford, Alec and Kim, Jong Wook and Hallacy, Chris and Ramesh, Aditya and Goh, Gabriel and Agarwal, Sandhini and Sastry, Girish and Askell, Amanda and Mishkin, Pamela and Clark, Jack and others},
  booktitle={ICML},
  pages={8748--8763},
  year={2021},
  organization={PmLR}
}

@article{hu2024storyagent,
  title={Storyagent: Customized storytelling video generation via multi-agent collaboration},
  author={Hu, Panwen and Jiang, Jin and Chen, Jianqi and Han, Mingfei and Liao, Shengcai and Chang, Xiaojun and Liang, Xiaodan},
  journal={arXiv preprint arXiv:2411.04925},
  year={2024}
}

@article{he2025dreamstory,
  title={Dreamstory: Open-domain story visualization by llm-guided multi-subject consistent diffusion},
  author={He, Huiguo and Yang, Huan and Tuo, Zixi and Zhou, Yuan and Wang, Qiuyue and Zhang, Yuhang and Liu, Zeyu and Huang, Wenhao and Chao, Hongyang and Yin, Jian},
  journal={TPAMI},
  year={2025},
  publisher={IEEE}
}

@article{zhang2023video,
  title={Video-llama: An instruction-tuned audio-visual language model for video understanding},
  author={Zhang, Hang and Li, Xin and Bing, Lidong},
  journal={arXiv preprint arXiv:2306.02858},
  year={2023}
}

@article{chen2023videollm,
  title={Videollm: Modeling video sequence with large language models},
  author={Chen, Guo and Zheng, Yin-Dong and Wang, Jiahao and Xu, Jilan and Huang, Yifei and Pan, Junting and Wang, Yi and Wang, Yali and Qiao, Yu and Lu, Tong and others},
  journal={arXiv preprint arXiv:2305.13292},
  year={2023}
}

@inproceedings{lin2024video,
  title={Video-llava: Learning united visual representation by alignment before projection},
  author={Lin, Bin and Ye, Yang and Zhu, Bin and Cui, Jiaxi and Ning, Munan and Jin, Peng and Yuan, Li},
  booktitle={EMNLP},
  pages={5971--5984},
  year={2024}
}

@misc{tan2025omnivideodemocratizingunifiedvideo,
      title={Omni-Video: Democratizing Unified Video Understanding and Generation}, 
      author={Zhiyu Tan and Hao Yang and Luozheng Qin and Jia Gong and Mengping Yang and Hao Li},
      year={2025},
      eprint={2507.06119},
      archivePrefix={arXiv},
      primaryClass={cs.CV},
      url={https://arxiv.org/abs/2507.06119}, 
}

@inproceedings{shi2024motion,
  title={Motion-i2v: Consistent and controllable image-to-video generation with explicit motion modeling},
  author={Shi, Xiaoyu and Huang, Zhaoyang and Wang, Fu-Yun and Bian, Weikang and Li, Dasong and Zhang, Yi and Zhang, Manyuan and Cheung, Ka Chun and See, Simon and Qin, Hongwei and others},
  booktitle={SIGGRAPH},
  pages={1--11},
  year={2024}
}

@inproceedings{yariv2025through,
  title={Through-the-mask: Mask-based motion trajectories for image-to-video generation},
  author={Yariv, Guy and Kirstain, Yuval and Zohar, Amit and Sheynin, Shelly and Taigman, Yaniv and Adi, Yossi and Benaim, Sagie and Polyak, Adam},
  booktitle={CVPR},
  pages={18198--18208},
  year={2025}
}

@inproceedings{li2026mask2iv,
  title={Mask2iv: Interaction-centric video generation via mask trajectories},
  author={Li, Gen and Zhao, Bo and Yang, Jianfei and Sevilla-Lara, Laura},
  booktitle={AAAI},
  volume={40},
  pages={6091--6099},
  year={2026}
}

@inproceedings{zhang2023adding,
  title={Adding conditional control to text-to-image diffusion models},
  author={Zhang, Lvmin and Rao, Anyi and Agrawala, Maneesh},
  booktitle={ICCV},
  pages={3836--3847},
  year={2023}
}

@article{wan2025wan,
  title={Wan: Open and Advanced Large-Scale Video Generative Models},
  author={Team Wan and Ang Wang and Baole Ai and Bin Wen and Chaojie Mao and Chen-Wei Xie and Di Chen and Feiwu Yu and Haiming Zhao and Jianxiao Yang and others},
  journal={arXiv preprint arXiv:2503.20314},
  year={2025}
}

@inproceedings{chen2023internvl,
  title={InternVL: Scaling up Vision Foundation Models and Aligning for Generic Visual-Linguistic Tasks},
  author={Chen, Zhe and Wu, Jiannan and Wang, Wenhai and Su, Weijie and Chen, Guo and Xing, Sen and Zhong, Muyan and Zhang, Qinglong and Zhu, Xizhou and Lu, Lewei and Li, Bin and Luo, Ping and Lu, Tong and Qiao, Yu and Dai, Jifeng},
  booktitle={Proceedings of the IEEE/CVF Conference on Computer Vision and Pattern Recognition (CVPR)},
  pages={24185--24198},
  year={2024}
}

\end{document}